\documentclass[lettersize,journal]{IEEEtran}
\usepackage{amsmath,amsfonts}
\usepackage{algorithmic}
\usepackage{algorithm}
\usepackage{array}
\usepackage[caption=false,font=normalsize,labelfont=rm,textfont=rm]{subfig}
\usepackage{textcomp}
\usepackage{stfloats}
\usepackage{url}
\usepackage{verbatim}
\usepackage{graphicx}
\usepackage[numbers,sort&compress]{natbib}
\usepackage{listings}
\usepackage[colorlinks=true,citecolor=blue,linkcolor=blue,urlcolor=blue]{hyperref}
\begin{document}

\title{Mahjax: A GPU-Accelerated Mahjong Simulator for Reinforcement Learning in JAX}

\author{
Soichiro~Nishimori\textsuperscript{1,2},
Shinri~Okano\textsuperscript{3},
Keigo~Habara\textsuperscript{4},
Sotetsu~Koyamada\textsuperscript{5,6,7},
Eason~Yu\textsuperscript{8},
and~Masashi~Sugiyama\textsuperscript{2,1}%
\thanks{
\textsuperscript{1} The University of Tokyo, Tokyo, Japan.
\textsuperscript{2} RIKEN AIP, Tokyo, Japan.
\textsuperscript{3} Nara Institute of Science and Technology, Nara, Japan.
\textsuperscript{4} Independent Researcher.
\textsuperscript{5} Kobe University, Kobe, Japan.
\textsuperscript{6} Kyoto University, Kyoto, Japan.
\textsuperscript{7} ATR, Kyoto, Japan.
\textsuperscript{8} The University of Sydney, Sydney, Australia.
Corresponding author: Soichiro Nishimori. Email: nishimori@ms.k.u-tokyo.ac.jp.
This work has been submitted to the IEEE for possible publication. Copyright may be transferred without notice, after which this version may no longer be accessible.
}
}



\maketitle

\bstctlcite{IEEEexample:BSTcontrol}

\begin{abstract}
    Riichi Mahjong is a multi-player, imperfect-information game characterized by stochasticity and high-dimensional state spaces.
    These attributes present a unique combination of challenges that mirror complex real-world decision-making problems in reinforcement learning.
    While prior research has heavily relied on supervised learning from human play logs to pre-train the policy, algorithms capable of learning \textit{tabula rasa} (from scratch) offer greater potential for general applicability, as evidenced by the AlphaZero lineage.
    To facilitate such research, we introduce \textbf{Mahjax}, a fully vectorized Riichi Mahjong environment implemented in JAX to enable large-scale rollout parallelization on Graphics Processing Units (GPUs).
    We also provide a high-quality visualization tool to streamline debugging and interaction with trained agents.
    Experimental results demonstrate that Mahjax achieves throughputs of up to \textbf{2 million} and \textbf{1 million steps per second} on eight NVIDIA A100 GPUs under the no-red and red rules, respectively.
    Furthermore, we validate the environment's utility for reinforcement learning by showing that agents can be trained effectively to improve their rank against baseline policies.
    The code is available at \url{https://github.com/nissymori/mahjax}.
\end{abstract}


\section{Introduction}\label{sec:introduction}

Riichi Mahjong is a popular tile-based game where players compete to form a winning hand under imperfect information \cite{li2020suphx}.
The game exemplifies complex real-world decision-making problems characterized by multi-agent interaction, high-dimensional state spaces, and stochasticity.
Consequently, it has been extensively studied in the field of reinforcement learning (RL) \cite{li2020suphx, zhao2022building, li2024tjong, ogami2024mj, fu2021actor}.

A significant milestone in this domain is Suphx \cite{li2020suphx}, the first AI to achieve top human-level performance in Mahjong.
While subsequent works have demonstrated strong results, they also predominantly rely on supervised learning (SL) from human logs \cite{li2020suphx} or offline RL \cite{Levine2020OfflineRL} for pre-training.
In contrast, the AlphaZero family of algorithms \cite{silver2016mastering, silver2017mastering, silver2018general} demonstrated that complex games can be mastered via \textit{tabula rasa} self-play without human priors.
This approach has recently extended to solving fundamental algorithmic problems \cite{fawzi2022discovering}.
Inspired by these achievements, solving Mahjong from scratch via pure RL remains a promising yet underexplored frontier.

However, self-play in complex environments necessitates a vast amount of trial-and-error experience.
For instance, AlphaHoldem \cite{zhao2022alphaholdem} required 6.5 billion training steps to master heads-up no-limit poker.
Given that Mahjong involves four players and longer horizons than poker, existing Central Processing Unit (CPU) based simulators create a computational bottleneck for practical training \cite{koyamada2022mjx}.
To address the data throughput challenge, the RL community has shifted toward hardware-accelerated environments \cite{koyamada2023pgx, bonnet2024jumanji, freeman2021brax, rutherford2024jaxmarl, matthews2024craftax, pignatelli2024navix}.
These vectorized environments enable agents to collect experience in massive batches directly on a Graphics Processing Unit (GPU), often yielding speedups exceeding $100\times$ over CPU baselines \cite{koyamada2023pgx, matthews2024craftax}.
Moreover, they facilitate novel algorithms that leverage massively parallel interactions \cite{gallici2024simplifying, macfarlane2024spo}.

Among existing frameworks, Pgx \cite{koyamada2023pgx} provides a suite of JAX-based board games but currently lacks a comprehensive implementation of complex imperfect-information games like Riichi Mahjong.
In this work, we introduce \textbf{Mahjax}, a fully vectorizable Riichi Mahjong environment written in JAX \cite{jax2018github}, designed to enable large-scale pure RL research.

Our contributions are summarized as follows:
\textbf{1) Vectorized Environment:} We provide a high-performance Mahjong environment adopting the Pgx Application Programming Interface (API), ensuring compatibility with modern JAX-based RL pipelines.
\textbf{2) Performance:} Mahjax scales efficiently across multiple GPUs, achieving up to \textbf{2 million} and \textbf{1 million steps per second} on eight NVIDIA A100 GPUs under the no-red and red rules, respectively.
\textbf{3) Usability:} We offer visualization tools to facilitate debugging and analysis.
\textbf{4) Validation:} We validate the environment through successful RL training, demonstrating its readiness for research.


\section{Related Work}\label{sec:related-work}
We review related work in the fields of Mahjong AI and GPU-accelerated RL environments.

\textbf{Mahjong in RL.}
Mahjong has been studied extensively in the RL literature.
Among agent-focused efforts, the most notable milestone is Suphx \citep{li2020suphx}, the first AI to achieve top human-level performance on Tenhou, the most popular Mahjong platform in Japan \citep{tsunoda2022tenhou}.
Since then, several agents have been developed by both commercial and open-source communities.
For example, NAGA\footnote{\url{https://dmv.nico/en/articles/mahjong_ai_naga/}}, developed by Dwango Media Village, reached the highest rank in Tenhou.
Mortal \citep{equim-chan2022mortal} serves as an open-source framework for training Mahjong agents.
A common feature of these works is that they employ SL or offline RL to pre-train the policy on human data collected from Tenhou, followed by fine-tuning via deep RL.
Several works have also explored variants of Mahjong.
For instance, \citet{zhao2022building} developed an agent for 3-player Mahjong (Sanma).
Additionally, \citet{ogami2024mj} proposed a method to improve player evaluation.

Regarding simulation infrastructure, Mjx \citep{koyamada2022mjx} offers a fast C++ simulator with a throughput of roughly 40k games per hour.
Similarly, Mortal provides a fast Rust-based simulator named Libriichi \citep{equim-chan2022mortal}, which achieves comparable speeds.
However, these CPU-based simulators face scalability limitations when attempting to leverage the massive parallelization required for large-scale self-play training.

\textbf{GPU-Accelerated Environments.}
Recently, there has been active development of environments written natively in JAX \citep{jax2018github} \citep{koyamada2023pgx, bonnet2024jumanji, freeman2021brax, rutherford2024jaxmarl, matthews2024craftax, pignatelli2024navix}.
Pgx \citep{koyamada2023pgx} provides classic board games like Go and Shogi, achieving speeds 10--100$\times$ faster than their CPU counterparts.
Other domains cover combinatorial optimization in Jumanji \citep{bonnet2024jumanji}, differentiable physics in Brax \citep{freeman2021brax}, and multi-agent tasks in JaxMARL \citep{rutherford2024jaxmarl}.
More recently, environments such as Craftax \citep{matthews2024craftax} for open-ended learning, Navix \citep{pignatelli2024navix} for grid-world navigation, and XLand-Minigrid \citep{nikulin2024xland} for meta-RL in grid-worlds have been introduced.
These vectorized environments not only accelerate simulation but also facilitate novel RL algorithms that leverage massively parallel interactions, such as parallel Q-learning (PQN) \citep{gallici2024simplifying}.
\section{Mahjax Overview}\label{sec:mahjax-overview}

\begin{figure}[t]
    \centering
    \includegraphics[width=0.50\textwidth]{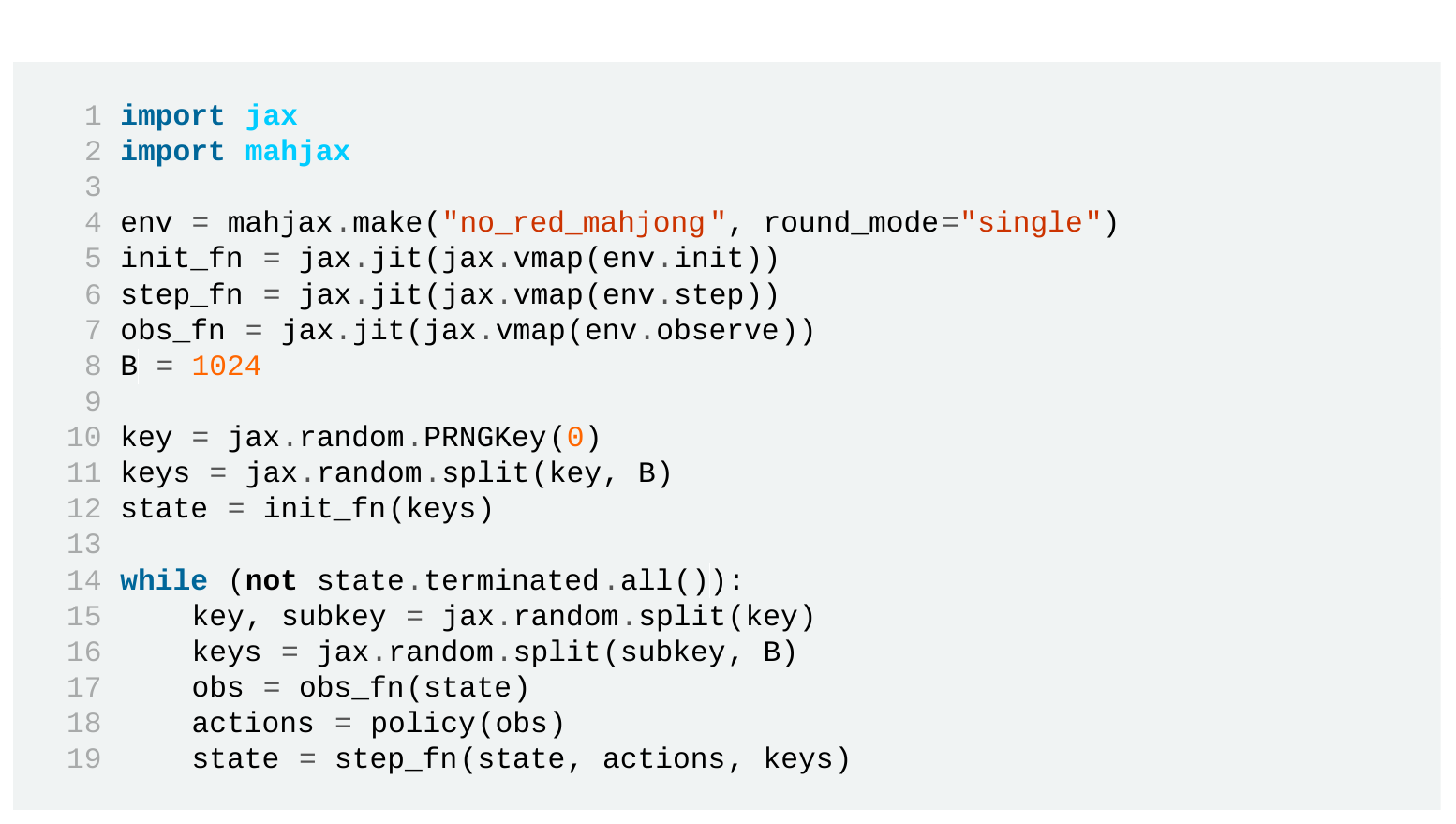}
    \caption{Example code snippet demonstrating the Mahjax API.}
    \label{fig:mahjax-code}
\end{figure}

In this section, we describe the design choices and implementation details of Mahjax.

\subsection{API Design and Implementation}
Mahjax adopts the API design of Pgx \citep{koyamada2023pgx} to ensure compatibility with fully vectorizable environments.
Figure \ref{fig:mahjax-code} illustrates a typical usage example.
To align with the JAX framework \citep{jax2018github}, we strictly adhere to a functional programming paradigm: the \texttt{State} dataclass stores all game information—including hands, scores, winds, melds, and masks—as immutable JAX arrays.
This design contrasts with prior Mahjong simulators that typically employ stateful, object-oriented architectures \citep{koyamada2022mjx}, which hinders the implementation in JAX.

Crucially, implementing game logic as pure functions is essential for JAX Just-In-Time (JIT) compilation.
However, Mahjong logic involves complex conditional branching, which can hinder parallel performance on GPUs.
To mitigate this, we employed two primary optimization techniques:
1) \textbf{Vectorized Logic:} We replaced control flow divergence (e.g., if-else statements) with matrix operations wherever feasible.
2) \textbf{Caching:} We implemented caching for computationally intensive evaluations, such as \textit{Yaku} (hand value) calculation.
Specifically, we pre-computed the relevant statistics for all possible suit combinations and encoded them into a bitmask.

\begin{figure}[t]
    \centering
    \includegraphics[width=0.48\textwidth]{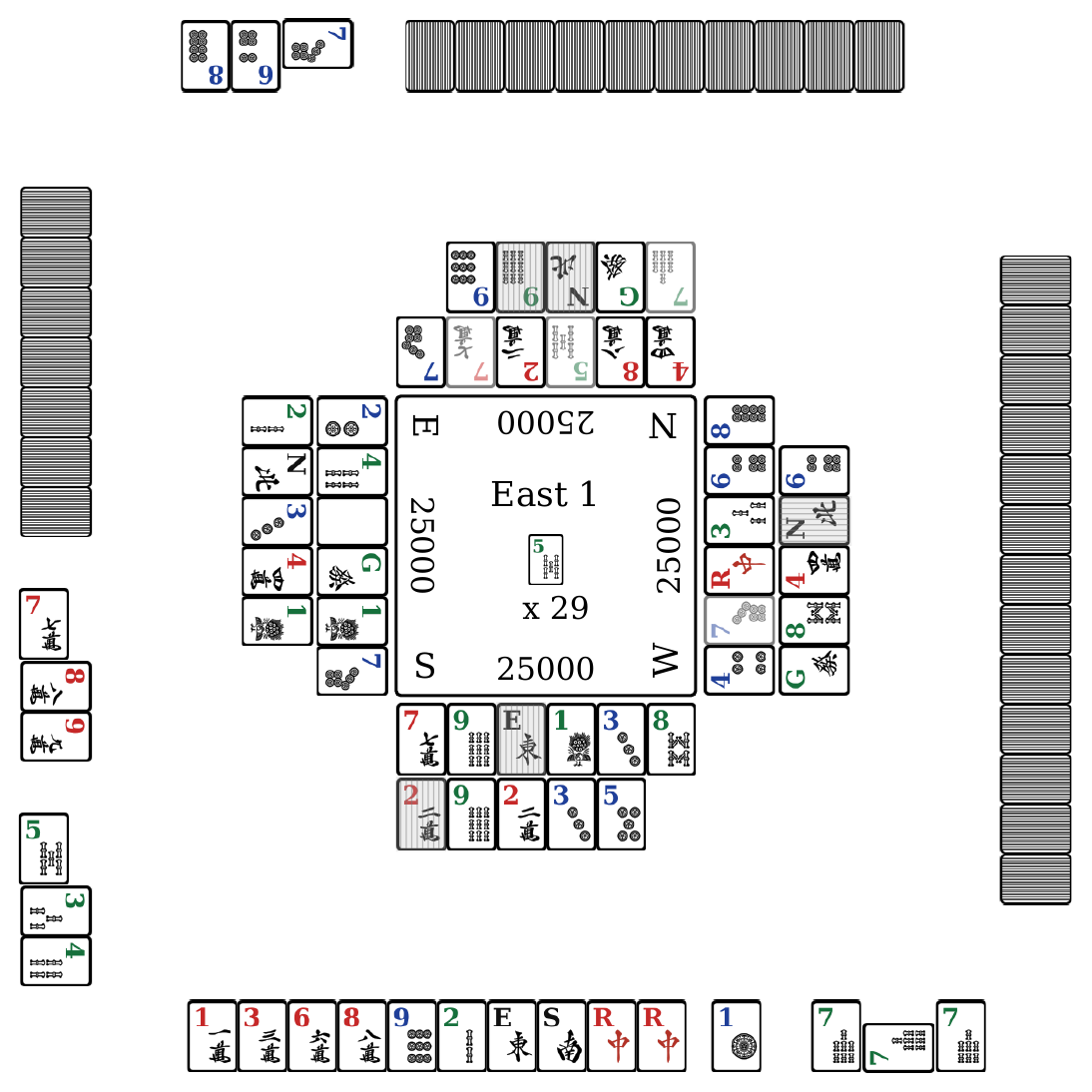}
    \caption{The SVG-based visualization of the Mahjax game state.}
    \label{fig:mahjax-visualization}
\end{figure}

\subsection{RL Environment Design}
Here, we describe the specific configurations of Mahjax as an RL environment.

\textbf{Rules.}
We adhere to the standard rules of four-player East-South \textit{Riichi} Mahjong.\footnote{\url{http://mahjong-europe.org/portal/images/docs/Riichi-rules-2025-EN.pdf}}
We support two major variants:
\begin{itemize}
    \item \textbf{Tenhou (Red) Rules:} The standard rules of four-player East-South \textit{Riichi} Mahjong as used in the Tenhou platform \citep{tsunoda2022tenhou}, including the red fives.
    Previous research has mainly focused on this variant \citep{li2020suphx,koyamada2022mjx,equim-chan2022mortal}.
    We validated the correctness of the implementation using downloaded play logs following \citet{koyamada2022mjx}.
    \item \textbf{No-Red Rules:} A variant of the game where red tiles are not used.
    For simplicity and higher throughput, we removed several complex rules such as abortive draw.
\end{itemize}

\textbf{Game Modes.}
To provide varying difficulty levels, we offer three modes: \texttt{single}, \texttt{east} and \texttt{half}.
In \texttt{single} mode, the episode terminates after a single \textit{Kyoku}, emphasizing immediate hand efficiency.
Conversely, \texttt{east} mode continues for up to 4 rounds (East-round only).
In \texttt{half} mode, the episode continues for up to 8 rounds (East and South), requiring long-term strategic planning, such as rank defense and temporary cooperation \citep{li2020suphx}.

\textbf{Action Space.}
The action space comprises discrete identifiers, covering discards, \textit{Kan}, and special moves (e.g., \textit{Riichi}, \textit{Ron}, \textit{Pon}, and \textit{Pass}).
A \texttt{legal\_action\_mask} is provided to filter invalid logits.
To enforce strict rule adherence, executing an illegal action triggers immediate termination with a penalty (default $-1.0$).

\textbf{Observation Space.}
Mahjax provides a structured dictionary observation for Transformer-based agents \citep{li2024tjong}.
It contains tokenized inputs such as hand indices, action history, and scalar properties (e.g., \textit{shanten number} and scores).
All observations are ego-centric to the current player.

\subsection{Visualization and UI}
Mahjax includes a Scalable Vector Graphics (SVG)-based visualization tool (Figure \ref{fig:mahjax-visualization}) and a web-based user interface.
These tools enable users to qualitatively analyze agent behaviors, debug the environment, and play against trained agents interactively.
To facilitate international research, the visualization supports English localization for users unfamiliar with traditional tile ideograms as shown in Figure \ref{fig:mahjax-visualization}.

\section{Experiments}\label{sec:experiment}
In this section, we evaluate the computational efficiency of Mahjax and validate its efficacy as a research platform for RL.

\subsection{Speed Benchmark}\label{sec:speed-benchmark}
\begin{figure}[t]
    \centering
    \subfloat[A100 x 1]{%
        \includegraphics[width=0.49\linewidth]{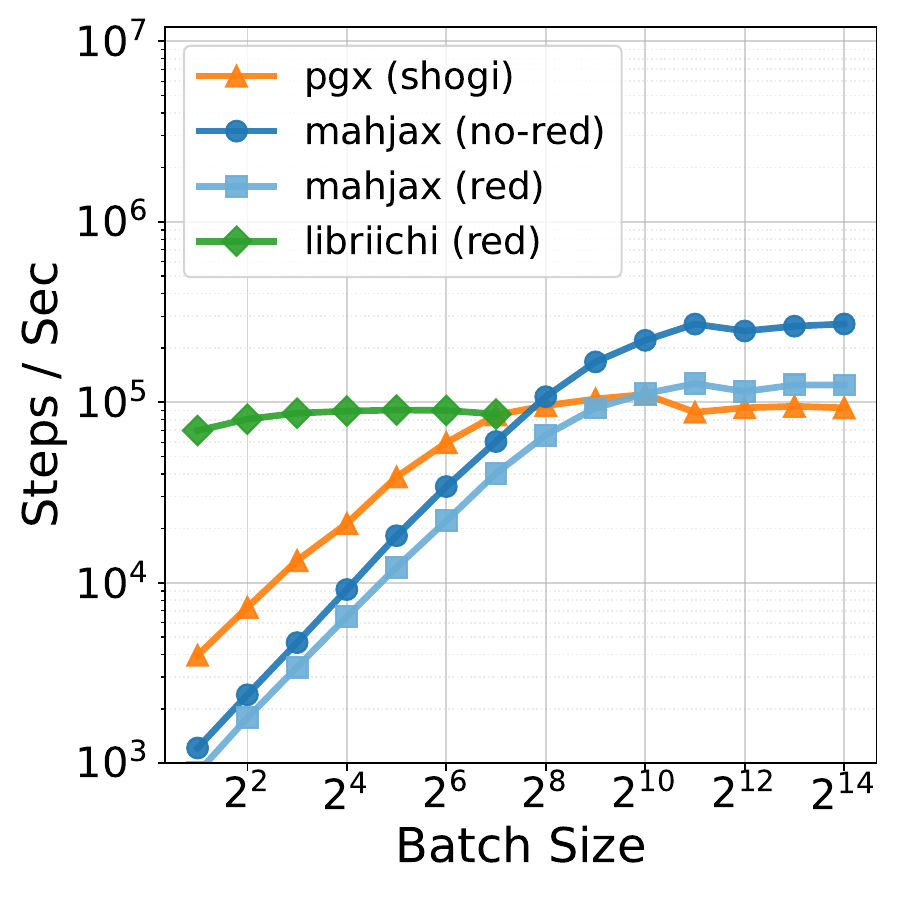}%
        \label{fig:speedbenchmark-1gpu}%
    }\hfill
    \subfloat[A100 x 8]{%
        \includegraphics[width=0.49\linewidth]{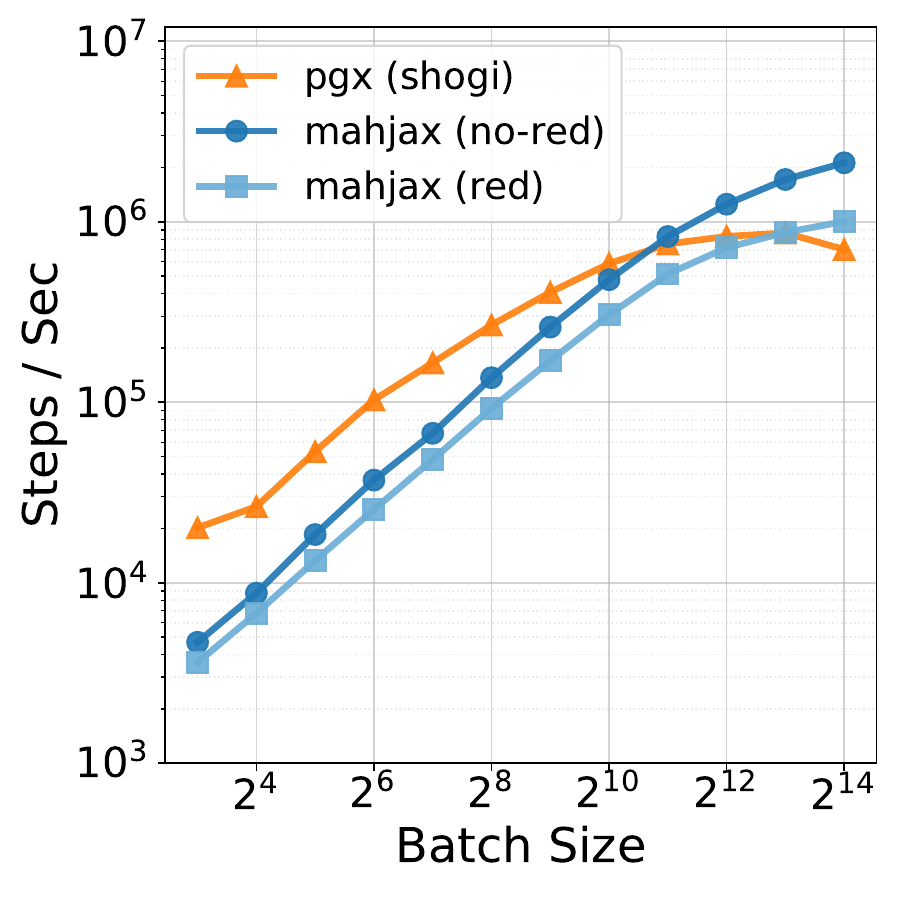}%
        \label{fig:speedbenchmark-8gpu}%
    }
    \caption{
        Throughput comparison (steps per second) between Mahjax (red and no-red rules), Pgx Shogi, and Libriichi across varying batch sizes.
        On a single GPU setting, Mahjax reaches a throughput plateau around batch size $2^{10}$.
        In contrast, on eight GPUs it continues to scale to larger batch sizes for both rule sets.
        Mahjax achieves peak throughputs of 2 million and 1 million steps per second on 8 GPUs for the no-red and red rules, respectively, outperforming Libriichi by over $10\times$ and surpassing Pgx Shogi.
    }
    \label{fig:speedbenchmark}
\end{figure}

\textbf{Setup.}
We compared Mahjax against two baselines:
1) \textbf{Libriichi} \citep{equim-chan2022mortal}, a Rust-based CPU simulator for the red-rule variant used in the Mortal project \citep{equim-chan2022mortal}; and
2) \textbf{Pgx (Shogi)} \citep{koyamada2023pgx}, the Shogi environment in Pgx.
In the absence of other GPU-accelerated Mahjong simulators, we included Pgx Shogi as a reference point to evaluate Mahjax's scalability.
Benchmarks were conducted on computing nodes equipped with two Intel Xeon Platinum 8360Y CPUs and eight NVIDIA A100 GPUs.
We measured throughput using \texttt{jax.pmap} for parallelization across devices for JAX environments, while utilizing Rayon \footnote{\url{https://github.com/rayon-rs/rayon}} for multi-threaded execution in Libriichi.
For Mahjax, we reported results for both the no-red and red-rule variants on a single GPU and eight GPUs to evaluate scalability.
Simulations ran for 100 batch steps using a random policy, with batch sizes ranging from 2 to 16,384 (starting from 8 for the 8-GPU setting).

\textbf{Results.}
Figure \ref{fig:speedbenchmark} shows the results.
On a single GPU, Mahjax scales with batch size up to roughly $2^{10}$ environments, after which throughput largely saturates, while Libriichi plateaus around $2^{3}$ due to CPU compute limitations.
In contrast, the 8-GPU configuration continues to scale beyond this regime for both rule variants, demonstrating effective multi-GPU parallelization.
Mahjax achieves peak throughputs of \textbf{2 million SPS} and \textbf{1 million SPS} on eight NVIDIA A100 GPUs for the no-red and red rules, respectively, outperforming Libriichi by over $10\times$ and surpassing Pgx Shogi.
These results confirm that Mahjax efficiently leverages GPU parallelism, making it suitable for large-scale training.

\begin{figure}[t]
    \centering
    \includegraphics[width=0.45\textwidth]{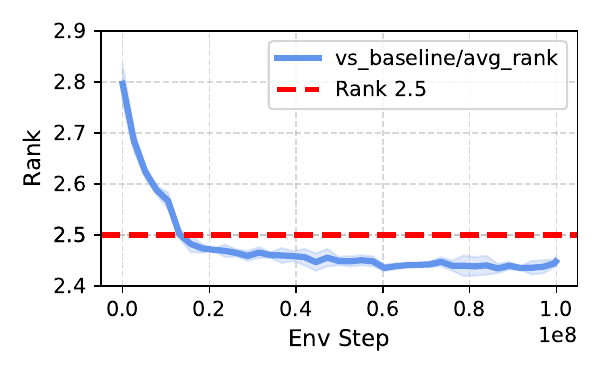}
    \caption{
        The plot shows the moving average rank against three fixed BC opponents over 1,000 evaluation games (lower is better).
        The solid line and shaded area represent the mean and standard deviation across three random seeds, respectively.
        The horizontal dotted line at 2.5 indicates the theoretical expected rank for players of equal skill.
    }
    \label{fig:rl-result}
\end{figure}

\subsection{RL Experiment}\label{sec:rl-experiment}
To validate the environment's stability for learning, we trained a policy using standard RL algorithms.

\textbf{Setup.}
We utilized the no-red rule and the \texttt{single-round} mode to accelerate experimental iteration.
To ensure training stability, we first initialized the policy via Behavioral Cloning (BC) \citep{li2020suphx} using 500k samples generated by a heuristic rule-based agent.
The agent architecture consists of a Transformer encoder \citep{vaswani2017attention} processing high-dimensional states (hand, discards, and other melds) into a latent representation, followed by separate multi-layer perceptron (MLP) heads for policy and value estimation.

Following BC pre-training, we fine-tuned the agent using Proximal Policy Optimization (PPO) \citep{schulman2017proximal} with Kullback-Leibler (KL) regularization towards the BC policy \citep{yu2025nash}.
Training employed 1,024 parallel environments with a rollout length of 256 steps.
Hyperparameters were set as follows: discount factor $\gamma=1.0$, Generalized Advantage Estimation parameter $\lambda=0.95$, learning rate $\eta=3\times10^{-4}$, clip range $\epsilon=0.2$, entropy coefficient $c_{\text{ent}}=0.01$, value coefficient $c_{\text{vf}}=0.5$, and KL penalty $c_{\text{KL}}=0.2$.
The training run spanned 100 million environmental steps, taking approximately 5.8 hours on a single NVIDIA GH200 Grace Hopper GPU.
To evaluate performance, we played the trained agent against three fixed BC policies (1 vs. 3) over 1,000 games.
Performance is measured by the average rank; since the expected rank for players of equal strength is 2.5, a lower value indicates superior performance.
We reported the average and standard deviation of three independent runs.

\textbf{Results.}
Figure \ref{fig:rl-result} presents the training trajectory.
The agent consistently achieves an average rank better (lower) than the neutral 2.5 baseline, indicating successful policy improvement over the BC initialization.
While aiming for state-of-the-art performance is beyond the scope of this paper, these results confirm that Mahjax provides a stable implementation for training deep RL agents.

\section{Concluding Remarks}\label{sec:conclusion}
In this work, we introduced \textbf{Mahjax}, a fully vectorizable Riichi Mahjong environment implemented in JAX.
Our experiments demonstrated that Mahjax achieves throughputs of up to \textbf{2 million} and \textbf{1 million steps per second} on eight NVIDIA A100 GPUs for the no-red and red rules, respectively, significantly outperforming the existing CPU-based simulator.
We further validated the environment's utility for RL by successfully training agents using PPO with KL regularization.

\textbf{Limitations and Future Work.}
Our current release supports both no-red and red-rule variants, but RL evaluation is still limited to the single-round setting and does not yet cover full round game mode.
Future work will expand rule support to include more game modes such as 3-player modes.
Furthermore, while our current RL training, leveraging BC pre-training, successfully demonstrates the simulator's reliability and meets the primary scope of this work, we aim to move toward learning from scratch.

\bibliographystyle{IEEEtranN}
{
\footnotesize
\bibliography{main}
}

\end{document}